# Study of visual processing techniques for dynamic speckles: a comparative analysis

Amit Chatterjee[*], Jitendra Dhanotiya[*], Vimal Bhatia[*] and Shashi Prakash[#]

[*] *Signals and Software Group, Discipline of Electrical Engineering, Indian Institute of Technology Indore, Indore-453446, India*

[#] *Photonics laboratory, Department of Electronics and Instrumentation Engineering, Institute of Engineering and Technology, Devi Ahilya University, Indore-452001, India*

Corresponding Author:

Shashi Prakash,

e-mail: sprakash_davv@rediffmail.com

tel: +91 731 2361116 7

fax: +91 731 2764385

Dynamic activity of a speckle pattern is a useful non-destructive tool to study various activities (e.g. ripening, drying, osmosis, Brownian motion, etc.) in biological samples. For analysis of dynamic speckle pattern, visual methods are preferred over numerical one because of their ability to generate full field 2D activity map. Main visual techniques used to obtain information from speckle patterns are Fujii's method, generalized difference (GD), weighted generalized difference, mean windowed difference (MWD), structural function (SF), modified structural function (MSF) etc. In this work, a comparative analysis of major visual techniques for natural gum sample collected from *Vachellia nilotica* tree is carried out. From obtained results it is observed that Fujii's method presents negative activity map due to its weighting coefficient. Although this problem is circumvented in GD and MWD based techniques; they suffer from high time complexity problem. In order to improve computational efficiency, SF and MSF methods have been proposed. These have time complexity in the order of Fujii's method along with a modified activity map. Activity parameters for all methods were estimated and tabulated for further comparison. Obtained results conclusively establish SF based method as an optimum tool for visual inspection of dynamic speckle data.

*Dynamic speckle, Fujii's method, GD, MWD, SF, MSF*

## Introduction

Dynamic speckle is non destructive, non invasive and non contact technique for analyzing activity of dynamic samples. When the sample is biological, the phenomenon is termed as 'bio-speckle'. In this method, a coherent source of light is used to illuminate desired sample. The backscattered lights interfere statistically; create a granular shaped speckle pattern [1]. Due to various physical and chemical activities within the living or non living dynamic sample, speckle patterns become mobile. For analyzing underlying activity of samples, dynamic-speckle images are recorded and processed through different image processing approaches. Processing techniques can be categorized into two broad areas namely global analysis and spatial analysis. Global analysis techniques [2] (inertia moment, absolute value of difference, wavelet entropy, autocorrelation, etc.) provide either numerical value corresponds to single column evolution of speckle frames or graphical results. Spatial analysis techniques (Fujii's method, alternative Fujii method, GD, modified GD, MWD, SF, MSF, etc.), also known as visual methods, provide full field two dimensional activity map as a result of pixel-by-pixel processing. For samples with different activity area (e.g. seed, animal tissue, leaf, etc.), spatial methods are preferred over numerical methods due to its capability of better perception of dynamic activity.

In this paper, we have analyzed and compared the performance of five most commonly used visual methods, i.e. Fujii's method, GD, MWD, SF and MSF. Sap of *Vachellia nilotica*, commonly known as Babool tree was used as a biological sample to conduct the experiment. For performance analysis, series of activity images correspond to initial and final stage of drying are collected using a combination of laser, beam expander and a high resolution CCD camera. In the next step, collected images are stacked together and processed using visual algorithms developed in MATLAB simulation environment. Two main parameters, i.e. 'mean activity difference' and 'code run time' are emphasized to compare the efficiency of spatial methods.

## Theory

In this section, definitions of visual descriptors are given for a sequence of images S=[ $I_0$, $I_1$, $I_2$, ..........,$I_N$], where $I_k$ (x, y) is the intensity level in the image k; k = 1, 2, .......,N for the pixel located in (x, y) coordinates, and N is the number of images in the acquired sequence.

1. Fujii's method: This descriptor was proposed by Fujii et al. [3] for processing bank of uniformly illuminated biospeckle images. This technique can be defined as a summation of absolute value difference between consecutive frames weighted by local average and mathematically given as,

$$\text{Fuji}(x, y) = \sum_{k=0}^{N-1} \frac{|I_k - I_{k+1}|}{I_k + I_{k+1}} \tag{1}$$

2. GD: This descriptor was proposed by Arizaga et al. [4] for uniformly as well as non-uniformly illuminated samples. It performs sum of absolute difference of values among all frames and mathematically given by,

$$\text{GD}(i, j) = \sum_{k=1}^{N} \sum_{l=k+1}^{N} |I_k(x, y) - I_l(x, y)| \tag{2}$$

where, k and l are indices that spans through all possible number of registered images.

3. MWD: This descriptor was proposed by Saude et al. [5] for increasing the sensitivity of algorithm by performing accumulation difference within a particular window length 'w'. This can be represented as:

$$\text{MWD}(i, j) = \sum_{k=1}^{N-w} \sum_{l=k+1}^{k+w} |I_k(x, y) - I_l(x, y)| \tag{3}$$

4. SF: This descriptor was proposed by Stoykova et al. [6] for uniform distribution of intensity fluctuation along the object. For a time lag $\tau = w\Delta t$, the estimate of temporal SF is determined mathematically as:

$$\text{SF}(i, j) = \sum_{k=1}^{N-w} (I_k(x, y) - I_{k+w}(x, y))^2 \tag{4}$$

5. MSF: This was proposed as an alternative to SF method by replacing the square term of Eq. 4 by absolute value of difference of $I_k(x,y)$-$I_{k+w}(x,y)$. This estimate is calculated as [7]:

$$MSF(i, j) = \sum_{k=1}^{N-w} \left| I_k(x, y) - I_{k+w}(x, y) \right| \qquad (5)$$

Eq. (4) and (5) contains single loop operation and therefore run faster than Eq. (2) and (3). With increase of activity, estimation values of Eq. (2)-(5) also increase while Eq. (1) may provide abrupt results due to presence of denominator term in the expression.

## Experimental Arrangement

For demonstration of comparative analysis, sap of *Vachellia nilotica* ( Babool tree), commonly known as 'Gum Arabic' or 'natural gum' has been collected in a glass tube as biological sample. A 15 mW red diode laser (λ=632.8 nm), source of coherent illumination was passed through a beam expander (40X) to increase field of view. Expanded beam, without any collimating lens then exposed to a coin surface coated with natural gum directly. Two sets of 30 speckle images of 300x400 pixels each were acquired using a CCD camera (BASLER, Germany, 1294x964 pixels resolution, 3.75 μm x3.75 μm pixel size and frame rate of 33 fps) at 10 minutes interval for visualization of gum drying process. Images are stored in host computer after digitizing to 8 bits by a frame grabber card and processed by image processing unit. A schematic of the experimental set up is shown in Fig. 1.

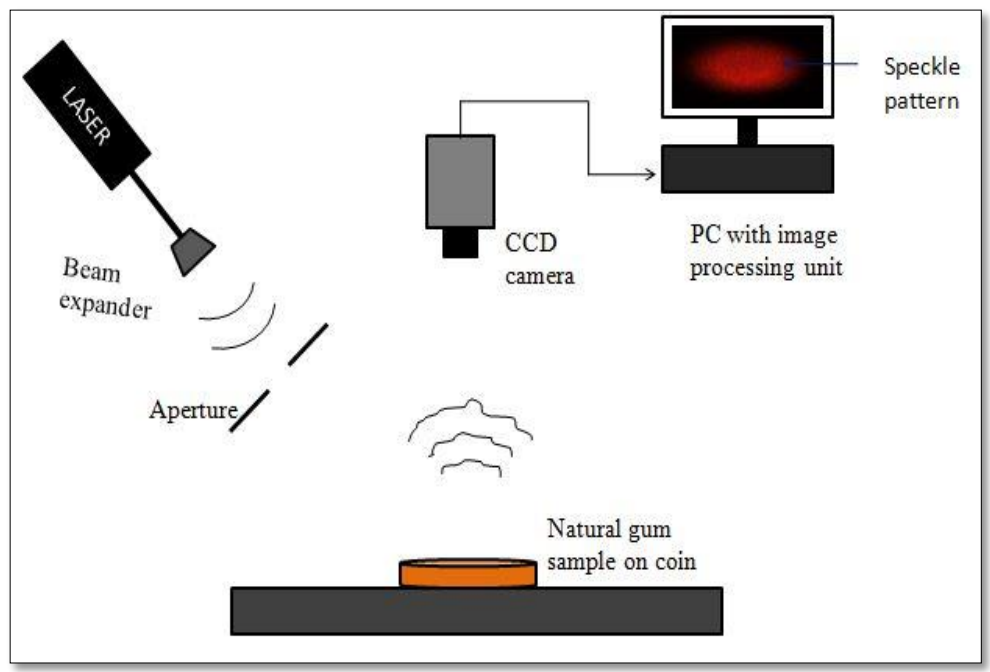

Fig. 1. Experimental setup for bio-speckle analysis of Gum Arabica drying.

## Results

For performing comparative study, algorithms are developed in MATLAB 7.0 with the system configuration: i3 processor, 4GB RAM, 4th generation and Windows 7 operating system. All the parameters correspond to high and low activity images of each method are calculated and listed in Table 1. Among them, two parameters, i.e. 'mean activity difference' (mean(X-X')) and 'code runtime' are considered for efficiency calculation and comparison among all methods.

In Fujii's method, averaging over reference gray levels produce false activity images as shown in Fig. 2. Here, a change in gray level from 0 to 1 produces an effect of 1 whereas a change from 254 to 255 produces 0.002. Although, in both the cases amount of change is 1, only change in reference  gray level leads to drastic changes in final Fujii image.

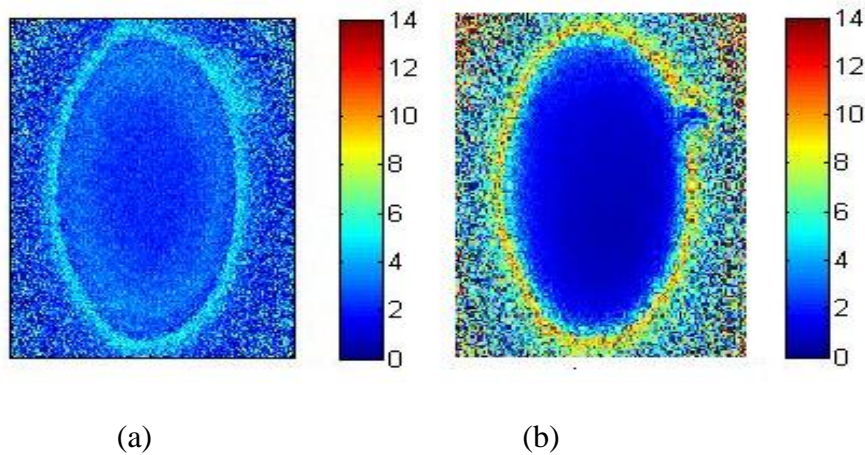

(a)                                      (b)

Fig. 2 Fujii images for natural gum sample at (a) t=0 minutes (high activity), and (b) t=10 minutes (low activity).

To circumvent this problem, GD method was introduced. In this method, the denominator term is casted and a 'sum of difference' is performed among all frames. A clear distinction between high and low activity is achieved as shown in Fig. 3. Although, this method possesses two serious drawbacks: firstly, presence of double loop makes this method computationally inefficient; secondly, even a small external noise can affect the results severely. MWD method was proposed as an alternative to conventional GD method in which the drawbacks of GD method was reduced considerably. The resultant images are shown in Fig. 4.

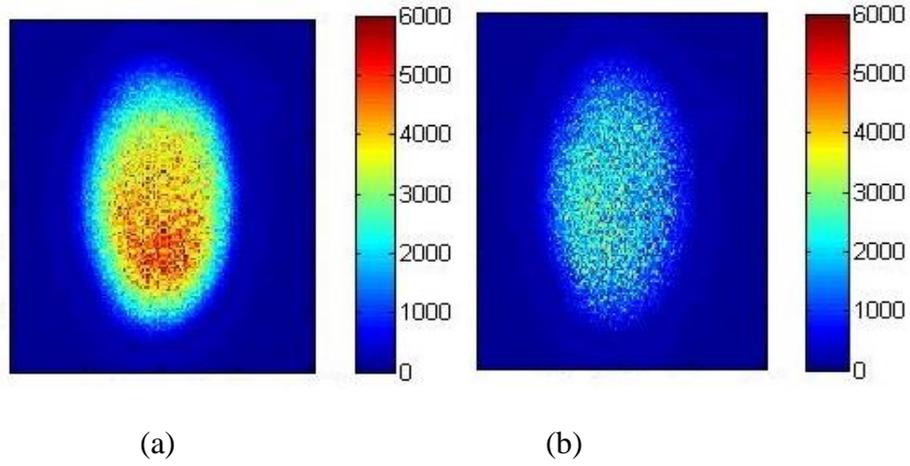

          (a)                  (b)

Fig. 3 GD images for natural gum sample at (a) t=0 minutes (high activity), and (b) t=10 minutes (low activity).

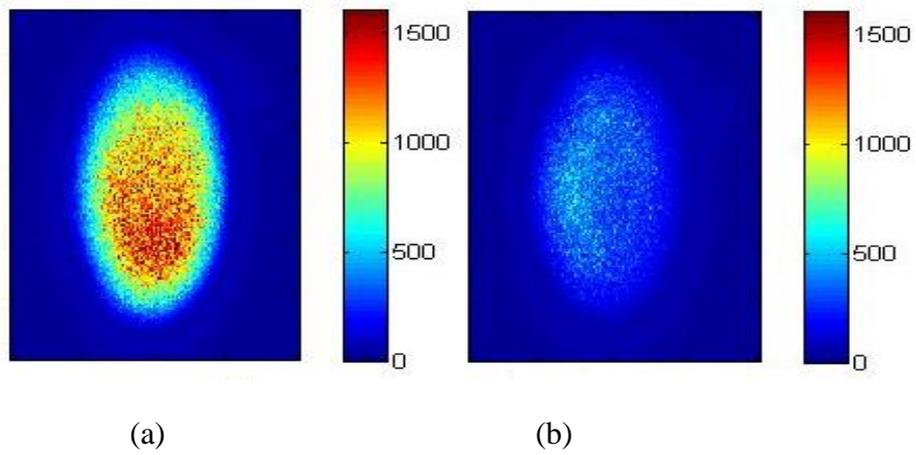

          (a)                  (b)

Fig. 4 MWD images for natural gum sample at (a) t=0 minutes (high activity), and (b) t=10 minutes (low activity).

However, with reduction of runtime, value of mean activity difference also reduced considerably in this technique.

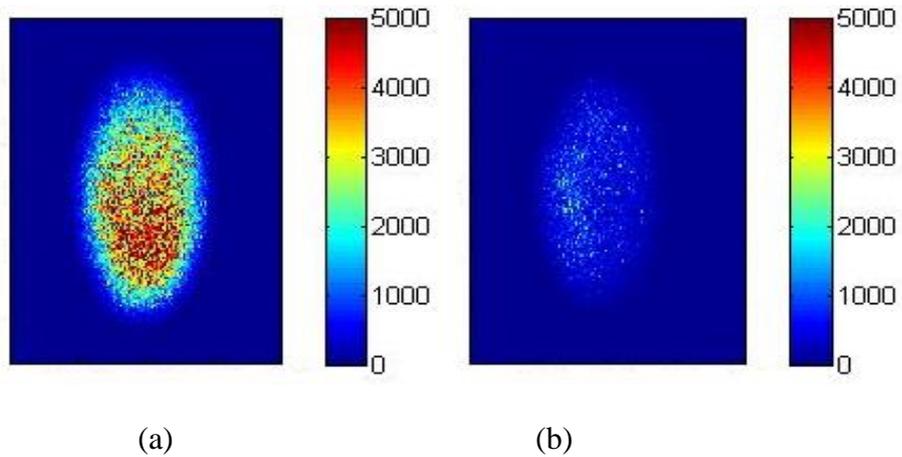

          (a)                  (b)

Fig. 5 SF images for natural gum sample at (a) t=0 minutes (high activity), and (b) t=10 minutes (low activity).

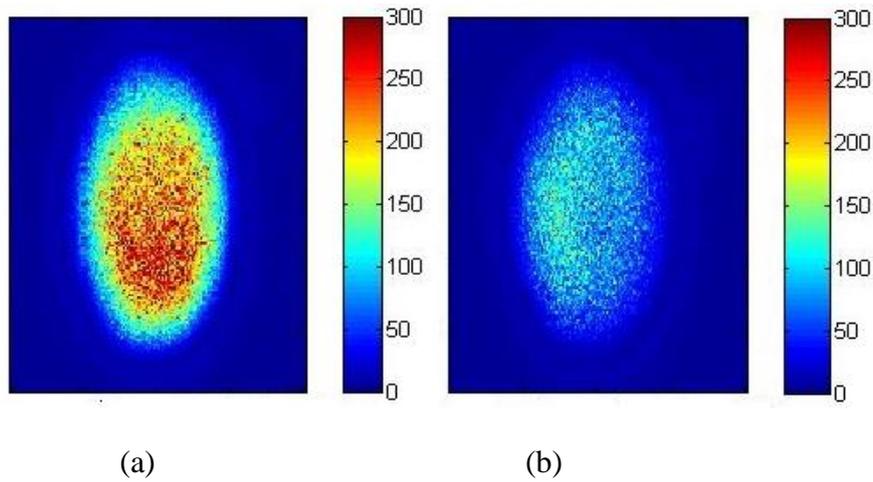

<center>(a)                                        (b)</center>

Fig. 6 MSF images for natural gum sample at (a) t=0 minutes (high activity), and (b) t=10 minutes (low activity).

Recently, two techniques namely SF and MSF were proposed as a fast alternative to conventional analysis methods (Fig. 5 and 6). In these methods, high computational speed is achieved by a combination of a single loop and window function. MSF used sum of absolute difference whereas SF used sum of squared difference for activity calculation. Due to presence of square term, SF amplifies the activity map. As a result, mean activity difference also increases in a higher extent to that of GD method.

| Image (I) | Max(I) (a.u.) | Min(I) (a.u.) | Mean(I) (a.u.) | Mean (X-X') (a.u.) | $t_{av}$=(t1*+t2*)/2 (seconds) |
|---|---|---|---|---|---|
| Fuji(X) | 2.8 | 0 | 3.05 | -1.3771 | 18.21 |
| Fuji(X') | 30.66 | 0 | 4.43 | | |
| MSF(X) | 485 | 0 | 55.15 | 29.77 | 13.59 |
| MSF(X') | 272 | 0 | 25.58 | | |
| SF(X) | 15946 | 0 | 674.01 | 560.96 | 13.33 |
| SF(X') | 4174 | 0 | 113.045 | | |
| GD(X) | 12225 | 0 | 999.15 | 441.8 | 252.66 |
| GD(X') | 8413 | 0 | 552 | | |
| MWD(X) | 2054 | 0 | 282.33 | 179.57 | 75.66 |
| MWD(X') | 901 | 0 | 102 | | |

Table 1 Numerical comparison of visual methods with window size w=5, where X= high activity and X'=low activity; t1* and t2* are calculated as an average of 20 distinct simulations.

## Discussion and Conclusion

In this paper, we have successfully demonstrated the comparative study of different visual inspection technique for dynamic speckle analysis. On consideration of mean activity difference as a parameter, GD and SF provides relatively better results amongst all discussed techniques, whereas SF and MSF provides better results while considering run time as the parameter. Conclusively, SF provides optimum measurement characteristics in concern of both parameters.

## Acknowledgement

This publication is an outcome of the R&D work undertaken project under the Visvesvaraya PhD Scheme of Ministry of Electronics & Information Technology, Government of India, being implemented by Digital India Corporation.